\documentclass[10pt,twocolumn,letterpaper]{article}

\usepackage{wacv}              

\usepackage{graphicx}
\usepackage{amsmath}
\usepackage{amssymb}
\usepackage{booktabs}
\usepackage{longtable}
\usepackage{makecell}
\usepackage{media9}
\usepackage{xcolor}

\usepackage[pagebackref,breaklinks,colorlinks]{hyperref}

\usepackage[capitalize]{cleveref}
\crefname{section}{Sec.}{Secs.}
\Crefname{section}{Section}{Sections}
\Crefname{table}{Table}{Tables}
\crefname{table}{Tab.}{Tabs.}

\def\wacvPaperID{30} 
\def\confName{WACV}
\def\confYear{2025}

\begin{document}

\title{SST-EM: Advanced Metrics for Evaluating Semantic, Spatial and Temporal Aspects in Video Editing}

\author{Varun Biyyala, Bharat Chanderprakash Kathuria, Jialu Li, and Youshan Zhang\\
Graduate Computer Science and Engineering Department\\
Katz School of Science and Health, Yeshiva University\\
205 Lexington Avenue, New York, NY 10016\\
{\tt\small \{biyyala, bkathuri, jli10\}@mail.yu.edu, youshan.zhang@yu.edu}}


\maketitle

\begin{abstract}
Video editing models have advanced significantly, but evaluating their performance remains challenging. Traditional metrics, such as CLIP text and image scores, often fall short: text scores are limited by inadequate training data and hierarchical dependencies, while image scores fail to assess temporal consistency. We present SST-EM (Semantic, Spatial, and Temporal Evaluation Metric), a novel evaluation framework that leverages modern Vision-Language Models (VLMs), Object Detection, and Temporal Consistency checks. SST-EM comprises four components: (1) semantic extraction from frames using a VLM, (2) primary object tracking with Object Detection, (3) focused object refinement via an LLM agent, and (4) temporal consistency assessment using a Vision Transformer (ViT). These components are integrated into a unified metric with weights derived from human evaluations and regression analysis. The name SST-EM reflects its focus on Semantic, Spatial, and Temporal aspects of video evaluation. SST-EM provides a comprehensive evaluation of semantic fidelity and temporal smoothness in video editing. The source code is available in the \textbf{\href{https://github.com/custommetrics-sst/SST_CustomEvaluationMetrics.git}{GitHub Repository}}.
\end{abstract}

\section{Introduction}
\label{sec: Introduction}

The rapid development of deep learning-powered video editing models has increased the need for effective evaluation methods. Traditional metrics like CLIP-based text and image similarity scores provide limited insights into model performance, particularly in dynamic video contexts. The CLIP text score, based on pre-trained Vision-Language Models (VLMs), struggles with generalization due to outdated or biased training data. The CLIP image score fails to account for temporal consistency, leading to inaccurate assessments when videos replicate original frames with minimal edits.

To address these limitations, we propose a novel multi-stage evaluation pipeline integrating multiple advanced models focusing on semantic fidelity, object presence, and temporal smoothness. Our pipeline operates in four stages: (1) semantic extraction using a VLM to compare frames with the editing prompt, (2) object tracking with Object Detection to ensure object consistency, (3) refinement through an LLM agent to focus on primary objects, and (4) temporal consistency evaluation using a Vision Transformer (ViT) model. We also introduce a unified metric that combines results from each stage, with weights determined through human evaluations and regression techniques.

Our proposed evaluation pipeline improves traditional methods by incorporating semantic understanding and temporal coherence. It demonstrates the effectiveness of combining VLMs, Object Detection, and Temporal Consistency checks, providing a more robust framework for video editing evaluation. Our empirically validated metric enables more reliable performance benchmarks in video editing research.

\begin{table*}[ht]
\centering
\caption{Comparison of Vision-Language Models (VLMs) and Object Detection Models.}
\resizebox{\textwidth}{!}{ 
\begin{tabular}{@{}lccccccc@{}}
\toprule
\textbf{Model} & \textbf{Vision-Language Tasks} & \textbf{Object Detection} & \textbf{Semantic Segmentation} & \textbf{Image Captioning} & \textbf{Zero-shot Learning} & \textbf{Backbone Architecture} & \textbf{Pretrained on Large-scale Dataset?} \\ \midrule
CLIP~\cite{clip}  & \checkmark & \checkmark & \checkmark & \checkmark & \checkmark & Vision Transformer (ViT) & \checkmark \\
Florence~\cite{florence}  & \checkmark & \checkmark & \checkmark & \checkmark & \checkmark & Swin Transformer & \checkmark \\
DETR~\cite{detr} & \textcolor{red}{$\times$} & \checkmark & \checkmark & \textcolor{red}{$\times$} & \textcolor{red}{$\times$} & Transformer & \checkmark \\
YOLOv7~\cite{yolov7} & \textcolor{red}{$\times$} & \checkmark & \textcolor{red}{$\times$} & \textcolor{red}{$\times$} & \checkmark & CNN-based (CSPDarknet) & \checkmark \\
ViLT~\cite{vilt} & \checkmark & \textcolor{red}{$\times$} & \textcolor{red}{$\times$} & \checkmark & \checkmark & Vision Transformer (ViT) & \checkmark \\
SAM~\cite{sam} & \checkmark & \checkmark & \checkmark & \checkmark & \checkmark & Vision Transformer (ViT) & \checkmark \\
MedCLIP~\cite{medclip} & \checkmark & \textcolor{red}{$\times$} & \checkmark & \checkmark & \checkmark & ViT & \checkmark \\
YOLOv5~\cite{yolov5} & \textcolor{red}{$\times$} & \checkmark & \textcolor{red}{$\times$} & \textcolor{red}{$\times$} & \checkmark & CNN-based (CSPDarknet) & \checkmark \\
Detectron2~\cite{detectron2} & \textcolor{red}{$\times$} & \checkmark & \checkmark & \textcolor{red}{$\times$} & \textcolor{red}{$\times$} & CNN-based (ResNet) & \checkmark \\
ActionCLIP~\cite{actionclip} & \checkmark & \textcolor{red}{$\times$} & \textcolor{red}{$\times$} & \checkmark & \checkmark & ViT & \checkmark \\
BLIP~\cite{blip} & \checkmark & \textcolor{red}{$\times$} & \checkmark & \checkmark & \checkmark & CNN-based (ResNet, ViT) & \checkmark \\
\bottomrule
\end{tabular}
}
\label{tab:vlm_object_detection_comparison}  
\end{table*}

\section{Related Work}


The evaluation of video editing models has been under-explored compared to other video-related tasks like video generation and action recognition. With the rise of deep learning-based video editing techniques, there has been a growing interest in developing robust and reliable metrics for evaluating video editing quality. However, most existing works primarily focus on image-based metrics or adapt video generation metrics, often overlooking the nuances of video editing.

\textbf{Evaluation Metrics for Video Editing Models.}
Historically, the evaluation of video editing models relied heavily on traditional image-based metrics, such as PSNR (Peak Signal-to-Noise Ratio), SSIM (Structural Similarity Index), and more recently, CLIP-based image and text similarity scores. These metrics provide valuable information about the visual similarity of individual frames or the relationship between video content and editing prompts. However, they are limited in capturing higher-order semantic relationships between objects and their temporal transitions, which are central to video editing tasks.

To address these limitations, some researchers have developed evaluation frameworks tailored to video editing. Kumar et al. \cite{kumar} proposed an evaluation system that combines semantic segmentation with temporal consistency to assess video content edits. Similarly, Zhou et al. \cite{zhou} explored the application of recurrent neural networks (RNNs) for capturing temporal dependencies between frames in edited videos, although their focus was limited to video generation and lacked the consideration of object-specific edits in video editing tasks.

\textbf{Object Detection in Video Editing and Generation.} Object detection plays a critical role in video editing, especially in ensuring that specific objects mentioned in an editing prompt are consistently modified across frames. Several works have explored object detection in the context of video generation. Carion et al. \cite{detr} introduced DETR (Detection Transformer), a powerful object detection model that integrates Transformer architectures for end-to-end object detection.

Recent advancements in object detection have also been incorporated into video analysis. Yolov7 \cite{yolov7} and DETR v2 have achieved state-of-the-art results in real-time object detection in videos, and their application to video editing tasks is promising. For instance, Chen et al. \cite{chen} proposed a framework where object detection models track object identities and their transformations over time, making it easier to evaluate how well an editing model maintains object consistency across frames. However, these models are typically evaluated with traditional metrics, such as Intersection over Union (IoU), which do not fully capture the importance of maintaining visual coherence and alignment with the editing prompt.

\textbf{Vision-Language Models (VLMs) for Video Editing Evaluation.} Integrating Vision-Language Models (VLMs) into video analysis has revolutionized how we evaluate video content. Models, such as CLIP (Contrastive Language-Image Pretraining), Florence2, and SAM (Segment Anything Model), excel at understanding visual content in the context of natural language, making them highly suitable for tasks that involve text-to-video or text-to-image relations. Godfrey et al. \cite{godfrey} demonstrated that CLIP can bridge the gap between visual and textual domains, providing a powerful metric for aligning visual content with text prompts. Florence2 and SAM extend this by providing more granular segmentation capabilities, enabling precise identification of objects and regions in images and videos.

Recent works, such as Lee et al. \cite{lee} and Dong et al. \cite{dong}, have applied these models to tasks like image captioning and scene understanding, which can be directly adapted to video editing evaluation by comparing the edited content to a given editing prompt. However, a challenge remains in integrating these models effectively with temporal consistency checks, as VLMs typically operate on a frame-by-frame basis without considering the dynamic nature of video content.

\textbf{Human Evaluation Scores in Video Editing.}
Human evaluation has long been regarded as the gold standard for assessing video editing quality due to its ability to capture subjective and nuanced aspects of edits that automated metrics often overlook. Human evaluators can assess various dimensions, such as semantic relevance to the editing prompt, visual quality, and temporal consistency, providing insights into the perceptual quality of the video edits.

Previous works have incorporated human evaluation scores to validate automated metrics in video editing. For instance, Xu et al. \cite{xu-HumanEval} conducted large-scale studies where participants scored videos on aspects like semantic fidelity and visual consistency. These scores were then used to benchmark automated metrics, revealing that traditional metrics like PSNR and SSIM often correlate poorly with human judgments in video editing tasks. Similarly, Wang et al. \cite{wang-humaneval} introduced a crowdsourced framework for evaluating text-to-video generation, where human evaluators scored the alignment of video content with given prompts. Their findings highlighted the importance of capturing higher-order semantic relationships and temporal coherence, which automated metrics struggle to emulate.

In the context of video editing, human evaluation also aids in understanding the relative performance of different models. For example, in recent studies by Zhao et al. \cite{zhao-humaneval}, human evaluations were used to compare various editing techniques on dimensions like artifact removal and natural transition consistency. However, the subjective nature of human evaluation introduces variability, making it crucial to design experiments with adequate inter-rater agreement metrics, such as Krippendorff’s alpha or Cohen’s kappa, to ensure reliability.

Despite its advantages, human evaluation has notable limitations, including its time-intensive nature, high costs, and dependence on subjective interpretations. These challenges have motivated researchers to use human scores as a benchmark to develop automated metrics that can approximate human judgments. To this end, regression techniques have been employed to optimize the weights of combined metrics against human evaluation scores, as seen in Kumar et al. \cite{kumar} and Zhou et al. \cite{zhou}.

\textbf{Combining Object Detection, VLMs, and Temporal Consistency.}
While some work integrated object detection and VLMs for video tasks, combining these with temporal consistency remains a relatively unexplored area. Pelechano et al. \cite{pelechano} introduce a framework that combines visual semantic understanding with temporal consistency models for video generation. Their approach focuses on maintaining a balance between frame-level object consistency and overall video coherence, while not fully exploring the integration of advanced models like ViT (Vision Transformers) for assessing temporal continuity between frames.

In this paper, we propose a novel pipeline that combines the strengths of VLMs for semantic analysis, Object Detection for object consistency, and Vision Transformers for temporal consistency, bridging the gap between these domains. Unlike previous work, our approach integrates all these elements to provide a more comprehensive evaluation of video editing models, accounting for both the accuracy of object edits and the smooth transitions between consecutive frames. We further leveraging human evaluation scores to fine-tune our proposed metric, \textbf{SST-EM}. Specifically, we regress our metric’s components against human evaluation results on one video editing model and validate its performance on another model. This approach ensures the metric aligns closely with human judgments while being robust to variations across different video editing techniques. Moreover, to demonstrate the generalizability of our metric, we evaluate its performance, alongside existing metrics, across four state-of-the-art video editing models using human evaluation as the baseline.

\section{Dataset Collection}

\begin{figure}[ht]
  \centering
  \includegraphics[width=\columnwidth]{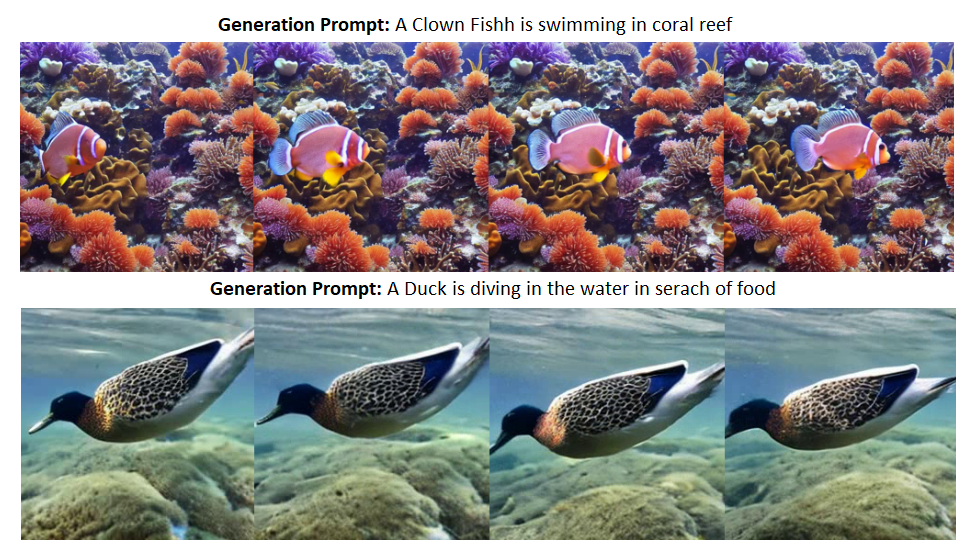}
  \caption{Examples of generated video frames with generation prompt.}
  \label{fig:Example Dataset Image}
\end{figure}

We curated a diverse dataset consisting of 40 distinct video pairs from the Enhanced End-to-End Video Editing dataset~\cite{enhanced}, which includes both original and edited video pairs (see Table~\ref{tab:dataset summary}). These videos were generated using a variety of state-of-the-art video editing models, such as MotionDirector\cite{zhao2025motiondirector}, Trailblazer\cite{trailblazer}, Tune-A-Video\cite{TuneAVideo}, and Text2LIVE\cite{text2live}. This extensive dataset allows us to comprehensively evaluate the video editing capabilities of different models across various tasks, ensuring a robust and diverse set of test cases (see Figure \ref{fig:Example Dataset Image}).

To further enhance the evaluation process, additional data from various other sources was collected to validate our mathematical evaluation formula. The collected dataset spans a wide range of motions, colors, and programmed zoom actions, designed to challenge the models in handling complex video synthesis scenarios. Each video sequence in this collection features a variety of motion dynamics, diverse background environments, and distinct trajectories. These videos provide a comprehensive set of sequences to assess the performance and robustness of the models in diverse editing contexts, including changes in movement patterns, lighting, and object manipulation.

\subsection{Data Preparation for Evaluation}
The dataset has been carefully organized into two distinct sets for optimizing and validating our evaluation model:

\textbf{Optimization Set}: This set is used to optimize the weights in our evaluation formula. The model is trained to fit the mathematical evaluation model using a linear regression methodology, with Human Evaluation (Human Eval) scores as the ground truth. Human Eval scores are subjective ratings collected from multiple individuals who were asked to assess the quality of the edited videos based on semantic accuracy, spatial coherence, and temporal consistency. To mitigate personal bias and differences in human perception, the evaluations were averaged across participants, ensuring a more reliable and objective evaluation process.

\textbf{Validation Set}: A separate set is used to validate the performance of our Mathematical Evaluation model. We compare the model's predictions with the Human Evaluation scores once again, ensuring the consistency and reliability of the evaluation methodology. This validation helps confirm our model's ability to replicate human judgment accurately in video editing evaluations.

For each video in the dataset, we convert the edited videos into individual frames, paired with their corresponding editing prompts. These frame-prompt pairs serve two purposes: optimizing the weights in our evaluation model and evaluating the performance of the video editing models. By maintaining this structure, we can efficiently assess how well the models perform across a wide range of editing tasks, from motion manipulation to visual style transfer.

\begin{table}[ht]
\centering
\caption{Summary of the Dataset}
\begin{tabular}{|p{2.5cm}|p{2cm}|p{2.5cm}|}
\hline
\textbf{Task Type} & \textbf{No. of Videos} & \textbf{Frame-Prompt Pairs} \\
\hline
Weights-Optimization & 40 & 640 \\
\hline
Evaluation & 40 & 900 \\
\hline
\end{tabular}
\label{tab:dataset summary}
\end{table}

\section{Methodology}

This section describes the detailed methodology underlying our SST benchmarking evaluation formula for the performance of video editing models. The evaluation process involves measuring several key aspects of the edited video, including semantic alignment, object detection accuracy, temporal consistency, and overall editing quality. These metrics are then combined into a final score using a weighted sum approach, where the weights are optimized through a linear regression model based on human evaluation scores.

\subsection{Stage 1: Context Similarity Score}

The Context Similarity Score \( S_{\text{similarity}} \) measures how closely the edited video matches the editing prompt in terms of semantic content. To compute this score, we use a Vision-Language Model (VLM), PaliGemma \cite{paligemma}, to generate textual captions for each video frame. PaliGemma is a state-of-the-art VLM known for its robust multimodal capabilities, enabling precise textual descriptions of visual content. It leverages extensive pretraining on diverse datasets, ensuring high accuracy in generating contextually rich and semantically aligned captions. This makes PaliGemma particularly suitable for evaluating video edits where understanding nuanced visual-text relationships is critical.

The similarity between the frame's caption \( \mathbf{C}_{\text{frame}} \) and the editing prompt \( \mathbf{C}_{\text{prompt}} \) is computed using the cosine similarity measure:

\[
S_{\text{similarity}} = \text{sim}(\mathbf{C}_{\text{frame}}\mathbf{C}_{\text{prompt}}),
\]
where \( \text{sim}(\cdot) \) denotes the cosine similarity function. The context similarity score quantifies how well the semantic content of the video matches the intended modifications described in the editing prompt.

\begin{figure}[ht]
  \centering
  \includegraphics[width=\columnwidth]{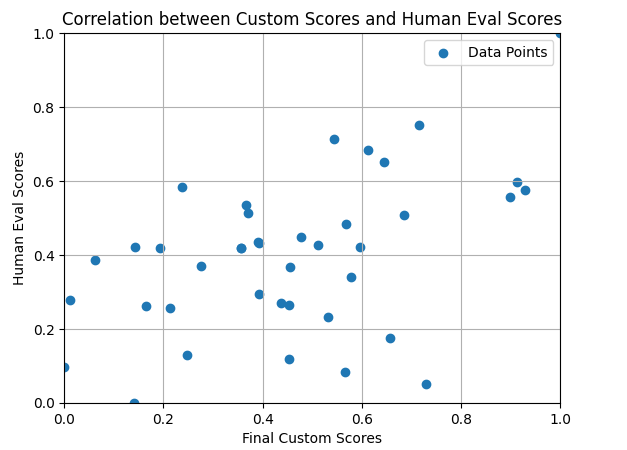}
  \caption{Comparison between Final and Human Eval Results}
  \label{fig:Comapare_results_Image}
\end{figure}

\subsection{Stage 2: Object Detection Score }

The Object Detection Score \( S_{\text{object\_detection}} \) evaluates how accurately the primary object described in the editing prompt is detected in each frame of the video. To achieve this, we utilize an object detection model that outputs confidence probabilities for detecting the specified object in each frame. The object detection score is computed by averaging the confidence scores across all video frames.

We employ the Grounding DINO model for this stage, which excels at text-conditioned object detection. Grounding DINO \cite{groundingDINO} integrates a transformer-based architecture that enables robust object identification by directly leveraging textual prompts. This capability is particularly beneficial in our evaluation framework, allowing us to seamlessly match the editing prompt with the detected objects in the video frames. By using editing prompts as input, Grounding DINO provides confidence scores for the presence of the described objects, ensuring alignment with the editing task and enabling precise evaluation of object-specific edits. Its high accuracy and ability to handle text-to-object associations make it well-suited for our methodology.

\[
S_{\text{object\_detection}} = \frac{1}{N} \sum_{i=1}^{N} P_{\text{object}}^{(i)},
\]
where \( P_{\text{object}}^{(i)} \) is the confidence probability for detecting the primary object in frame \( i \), and \( N \) is the total number of frames in the video.

At this stage, we also use an LLM agent, Mistral-7B-Instruct-v0.3, to help focus the object detection model on the primary object specified in the editing prompt, filtering out any irrelevant objects or background noise. Using Prompt Engineering, a carefully crafted prompt dynamically takes the editing prompt and passes it to the LLM agent, which in turn outputs the primary object of interest from the editing prompt. This ensures that the object detection model is specifically guided to detect and track the relevant object throughout the video, improving the accuracy and relevance of the object detection score.


\subsection{Stage 3: Temporal Consistency Score }

The Temporal Consistency Score \( S_{\text{temporal}} \) measures the smoothness and coherence of transitions between consecutive frames in the video. A key aspect of video editing is ensuring that changes are smoothly applied over time, maintaining visual consistency. To calculate the temporal consistency score, we use a Vision Transformer (ViT) model, which computes embeddings for each frame in the video. The similarity between consecutive frame embeddings is measured using cosine similarity. The temporal consistency score is defined as:

\[
S_{\text{temporal}} = \frac{1}{N-1} \sum_{i=1}^{N-1} \text{sim}(\mathbf{F}_i, \mathbf{F}_{i+1}),
\]
where \( \mathbf{F}_i \) and \( \mathbf{F}_{i+1} \) are the feature embeddings for consecutive frames \( i \) and \( i+1 \), and \( N \) is the total number of frames in the video. Higher values of \( S_{\text{temporal}} \) indicate better temporal consistency, where transitions between frames are smoother.

\subsection{Stage 4: Final Score Calculation}

\begin{figure*}[ht]
  \centering
  \includegraphics[width=\textwidth]{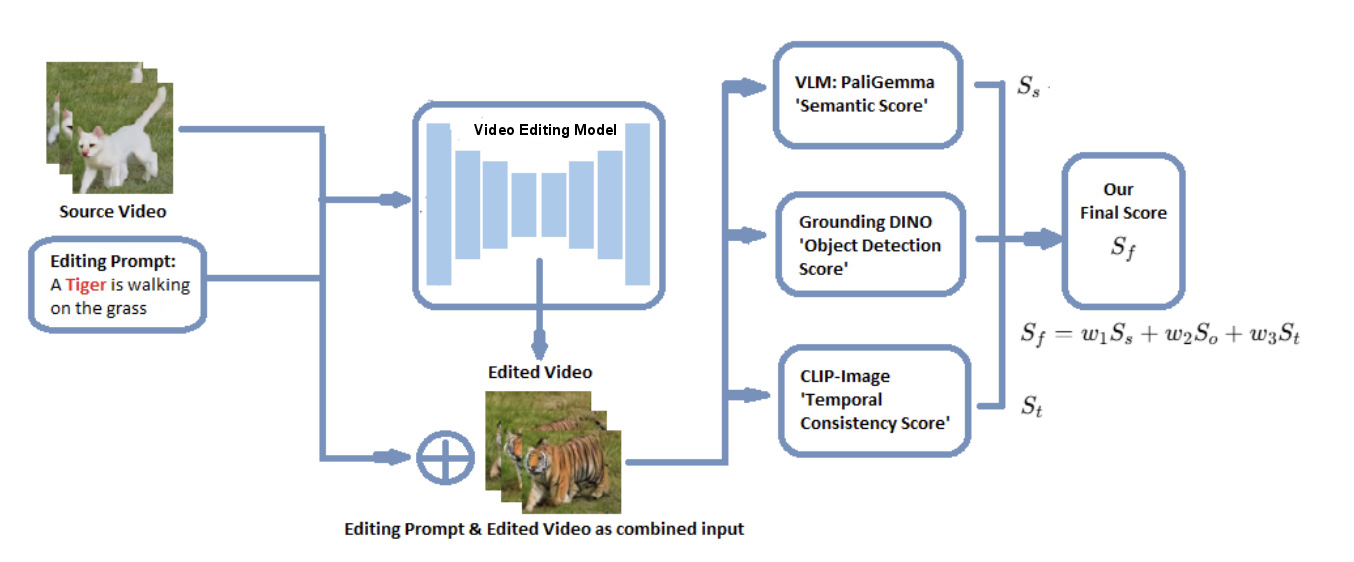}
  \vspace{-0.6cm}
  \caption{Pipeline for evaluating video editing models using custom metrics: semantic analysis, object detection, and temporal consistency}
  \label{fig:evaluation_metrics_pipeline}
\end{figure*}

The final evaluation score \( S_{\text{final}} \) is computed as a weighted sum of the three metrics: context similarity, object detection, and temporal consistency. The final score is given by:

\[
S_{\text{final}} = w_1 \cdot S_{\text{similarity}} + w_2 \cdot S_{\text{object\_detection}} + w_3 \cdot (S_{1-\text{temporal}}),
\]
where \( w_1 \), \( w_2 \), and \( w_3 \) are the weights corresponding to each metric, and \( S_{\text{temporal}} \) is subtracted from 1 to ensure that a higher temporal consistency score yields a higher final score (see Figure \ref{fig:evaluation_metrics_pipeline}).

\subsection{Optimization and Regression}

The weights \( w_1 \), \( w_2 \), and \( w_3 \) for the final score formula are optimized using a linear regression model. 

To derive these weights, we first organize the data into two distinct sets:

\textbf{Optimization Set}: Used to optimize the weights by fitting the mathematical evaluation model to human evaluation scores. Human scores are gathered from multiple evaluators to mitigate individual biases and perception differences. The objective is to minimize the difference between the predicted scores from our model and the human evaluation scores.

\textbf{Validation Set}: Used to validate the performance and consistency of the mathematical evaluation model. The model's predictions are compared against human evaluation scores to ensure consistent behavior.

Each edited video is converted into frames, and each frame is paired with its corresponding editing prompt. These pairs are then used to compute three individual scores (context similarity, object detection, and temporal consistency) for each video. These calculated scores are subsequently used as input features for the linear regression model training.

The optimization objective is to minimize the error between the predicted final scores and the human evaluation scores. This is done by solving the following loss function:

\[
\mathcal{L} = \frac{1}{M} \sum_{j=1}^{M} \left( S_{\text{final}}^{(j)} - S_{\text{human}}^{(j)} \right)^2,
\]
where \( S_{\text{final}}^{(j)} \) is the final score predicted by the model for the \( j \)-th video, and \( S_{\text{human}}^{(j)} \) is the corresponding human evaluation score. \( M \) is the number of videos in the dataset used for training.

The optimization set is used to optimize the weights \( w_1 \), \( w_2 \), and \( w_3 \), ensuring that the model accurately reflects human judgment. The validation set is then used to test the generalization ability of the model and ensure consistent evaluation behavior across different videos and editing tasks.

\subsection{Finalizing Weights Optimization}

Once the initial optimization process is complete, we compare the results generated from the final score formula with the human evaluation results to assess the consistency of our model. This is done using statistical methods such as correlation and the R1 score, which are commonly used to measure the degree of similarity between predicted and actual values.

We calculate the Pearson correlation coefficient \( \rho \) between the predicted final scores and the human evaluation scores:

\[
\rho = \frac{\sum_{i=1}^{M} (S_{\text{final}}^{(i)} - \overline{S_{\text{final}}}) (S_{\text{human}}^{(i)} - \overline{S_{\text{human}}})}{\sqrt{\sum_{i=1}^{M} (S_{\text{final}}^{(i)} - \overline{S_{\text{final}}})^2 \sum_{i=1}^{M} (S_{\text{human}}^{(i)} - \overline{S_{\text{human}}})^2}},
\]
where \( \overline{S_{\text{final}}} \) and \( \overline{S_{\text{human}}} \) are the mean predicted and human evaluation scores, respectively. A higher correlation indicates that the model can predict human evaluation scores accurately (see Figure~\ref{fig:Comapare_results_Image}).

Additionally, we calculate the R1 score, which measures the precision of the predicted final scores in terms of their alignment with human evaluations. High correlation and R1 scores indicate that the model's predictions are consistent with human judgment, confirming that the weight optimization process has been successful.

By comparing the optimized final scores with human evaluations through these metrics, we ensure that the final model is robust and produces reliable results that align with human perception.

\begin{figure*}[ht]
  \centering
  \includegraphics[width=\textwidth]{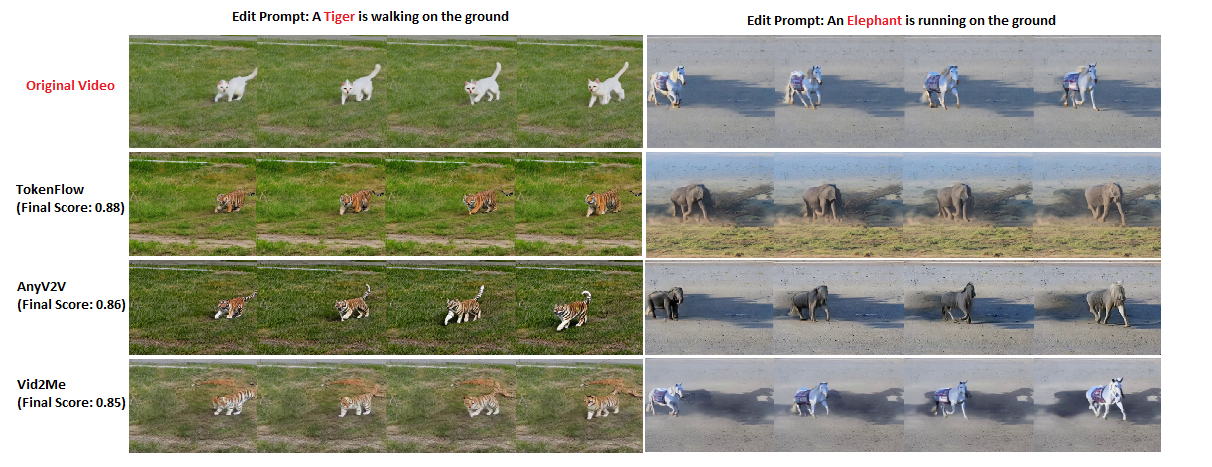}
  \vspace{-0.3cm}
  \caption{Comparison of results between different video editing models. The primary object of interest and the original video are highlighted in red.}
  \label{fig:Edited_results_comparison_models}
\end{figure*}

\section{Results}

We compare the performance of our SST-EM evaluation framework with various established metrics using video editing models such as VideoP2P, TokenFlow, Control-A-Video, and FateZero. Video results are displayed in Figure \ref{fig:Edited_results_comparison_models}. Correlation analysis (Pearson, Spearman, and Kendall) is conducted to compare all available metrics, including SST-EM, which is based on the Human Evaluation Score. The Human Evaluation Score integrates key aspects such as Imaging Quality, FF-$\alpha$, FF-$\beta$, Background Consistency Score, Success Rate, Subject Consistency, and Aesthetic Quality.

The human evaluation assesses critical aspects like semantic alignment, object detection accuracy, temporal consistency, and overall editing quality. These are combined into a final score using a weighted sum approach.

An ablation study is conducted on video editing models like AnyV2V, Vid2Me, VideoP2P, TokenFlow, Control-A-Video, and FateZero, with each model evaluated across Semantic Similarity Score, Object Detection Score, Temporal Consistency Score, SST-EM final score, and the CLIP-Text score (see Table \ref{tab:comparison of segments of custom metrics}).

Table \ref{tab:comparison between different metrics} compares models based on Imaging Quality, FF-$\alpha$, FF-$\beta$, Background Consistency, Success Rate, Subject Consistency, Aesthetic Quality, and SST-EM final score using Human Evaluation Score. This table offers a detailed breakdown of each model’s performance and demonstrates why SST-EM is a superior framework for evaluating video editing.

These results highlight SST-EM's advantages in capturing overall video editing quality and its correlation with human judgment.

\subsection{Comparison with Other Metrics}
We computed Pearson, Spearman, and Kendall correlations between SST-EM and other evaluation metrics. The results, summarized in Table \ref{tab:corelation_comparison_metrics}, show the alignment between our final score and other metrics assessing video quality. The individual components of SST-EM, Context Similarity Score, Object Detection Score, Temporal Consistency Score also show notable correlation with the Human Evaluation Score, demonstrating the contribution of each component to the final score.

\begin{table}[ht]
\centering
\caption{Comparison of Correlations of all the metrics with Human Evaluation Scores. Our final score and individual component scores are highlighted}
\vspace{-0.3cm}
\resizebox{\columnwidth}{!}{%
\begin{tabular}{|c|c|c|c|}
\hline
\textbf{Metric} & \textbf{Pearson} & \textbf{Spearman} & \textbf{Kendall} \\ \hline
Imaging Quality~\cite{chen2024editboard} & 0.951 & 0.800 & 0.666 \\ \hline
FF-$\alpha$~\cite{chen2024editboard} & -0.515 & -0.800 & -0.666 \\ \hline
FF-$\beta$~\cite{chen2024editboard} & -0.652 & -0.800 & -0.666 \\ \hline
Background Consistency Score~\cite{chen2024editboard} & 0.724 & -0.600 & -0.333 \\ \hline
Success Rate~\cite{chen2024editboard} & 0.794 & 0.800 & 0.666 \\ \hline
Subject Consistency~\cite{chen2024editboard} & 0.827 & 0.800 & 0.666 \\ \hline
Aesthetic Quality~\cite{chen2024editboard} & 0.837 & 0.946 & 0.912 \\ \hline
\textbf{Context Similarity Score} & 0.072 & -0.400 & -0.333 \\ \hline
\textbf{Object Detection Score} & 0.835 & 0.800 & 0.666 \\ \hline
\textbf{Temporal Consistency Score} & 0.927 & 1.000 & 1.000 \\ \hline
\textbf{SST-EM Score} & 0.962 & 1.000 & 1.000 \\ \hline
\end{tabular}%
}
\label{tab:corelation_comparison_metrics}
\end{table}
\begin{table*}[ht]
\centering
\caption{Comparison between different elements of our Framework}
\vspace{-0.3cm}
\resizebox{\textwidth}{!}{%
\begin{tabular}{lcccccc}
\toprule
\textbf{Model Name} & \textbf{CLIP-Text} & \textbf{Semantic Similarity Score} & \textbf{Object Detection Score} & \textbf{Temporal Consistency Score} &  \textbf{*Our SST-EM Score} \\
\midrule
AnyV2V\cite{anyv2v}           & 0.2918 & 0.780449 & 0.719760 & 0.966250 & \textbf{0.864687} \\
Vid2Me\cite{vid2me}           & 0.2747 & 0.725520 & 0.724870 & 0.978866 & \textbf{0.851886} \\
VideoP2P\cite{videop2p}         & 0.2783 & 0.749997 & 0.701194 & 0.932500 & \textbf{0.834226} \\
TokenFlow\cite{tokenflow}        & 0.2915 & 0.794004 & 0.743507 & 0.979918 & \textbf{0.879671} \\
Control-A-Video\cite{controlavideo}  & 0.2764 & 0.732529 & 0.738756 & 0.958416 & \textbf{0.846039} \\
FateZero\cite{fatezero}         & 0.3137 & 0.729603 & 0.736148 & 0.972566 & \textbf{0.851731} \\
\bottomrule
\end{tabular}%
}
\label{tab:comparison of segments of custom metrics}
\end{table*}

\begin{table*}[ht]
\centering
\caption{Comparison of our Framework with different other metrics using Human Evaluation Score}
\vspace{-0.3cm}
\resizebox{\textwidth}{!}{%
\begin{tabular}{lccccccccc}
\toprule
\textbf{Model Name} & \textbf{Imaging Quality} & \textbf{FF-}\(\boldsymbol{\alpha}\)
 & \textbf{FF-}\(\boldsymbol{\beta}\)
 & \makecell{\textbf{Background} \\ \textbf{Consistency Score}} & \makecell{\textbf{Success} \\ \textbf{Rate}} & \makecell{\textbf{Subject} \\ \textbf{Consistency}} & \makecell{\textbf{Aesthetic} \\ \textbf{Quality}} & \makecell{\textbf{Human} \\ \textbf{Evaluation Score}} &\makecell{\textbf{*Our} \\ \textbf{SST-EM Score}} \\
\midrule
VideoP2P\cite{videop2p}         & 0.6665 & 11.8893 & 0.2216 & 0.9696 & 0.5156 & 0.9692 & 0.4847& 0.411 & \textbf{0.834226} \\
TokenFlow\cite{tokenflow}        & 0.7408 & 7.2708  & 0.1566 & 0.9525 & 0.6471 & 0.9790 & 0.5546 & 0.452& \textbf{0.879671} \\
Control-A-Video\cite{controlavideo}  & 0.6973 & 18.0534 & 0.2674 & 0.9700 & 0.6050 & 0.9672 & 0.5377 & 0.425& \textbf{0.846039} \\
FateZero\cite{fatezero}         & 0.6907 & 8.0082  & 0.1723 & 0.9497 & 0.5294 & 0.9696 & 0.5546 & 0.433& \textbf{0.851731} \\
\bottomrule
\end{tabular}%
}
\label{tab:comparison between different metrics}
\end{table*}


\subsection{Pearson Correlation}

The Pearson correlation coefficient \( \rho \) measures the linear relationship between two variables, providing a value between -1 and +1. A value of +1 indicates a perfect positive linear relationship, -1 indicates a perfect negative linear relationship, and 0 indicates no linear relationship.

In our evaluation, Imaging Quality and Aesthetic Quality show high Pearson correlations with Human Evaluation scores (0.951 and 0.837, respectively), indicating strong linear relationships (see Table \ref{tab:corelation_comparison_metrics}). However, our SST-EM  score surpasses all other metrics with a Pearson correlation of 0.962. Moreover, the individual components of SST-EM, particularly the Temporal Consistency and Object Detection Scores, demonstrate Pearson correlations of 0.927 and 0.835, respectively, highlighting their significance in evaluating video editing quality.

\subsection{Spearman Rank Correlation}

The Spearman rank correlation assesses how well the rankings of two variables agree. The Spearman's correlation considers the relative order of values. It is useful when the relationship between variables is monotonic but not necessarily linear.

A Spearman correlation ranges from -1 to +1, with +1 indicating perfect agreement between rankings, and -1 indicating a perfect disagreement. Our SST-EM final score achieves a Spearman correlation of 1.000 with both Imaging Quality and Aesthetic Quality, demonstrating perfect rank agreement. 

\subsection{Kendall Tau Correlation}

The Kendall Tau correlation coefficient measures the strength of association between two variables based on the concordance and discordance of their pairs.

Kendall Tau values range from -1 to +1, with +1 indicating perfect agreement, -1 indicating perfect disagreement, and 0 indicating no agreement. Our SST-EM final score again shows a perfect agreement with Imaging Quality and Aesthetic Quality, yielding a Kendall Tau of 1.000. 

\subsection{Interpretation of Results}

Table~\ref{tab:corelation_comparison_metrics} highlights the performance of each metric compared to our SST-EM final score. Imaging Quality and Aesthetic Quality exhibit strong positive correlations with the Human Evaluation score, but our SST-EM surpassed all scores, particularly in Pearson and Spearman correlations. This indicates that our final score aligns well with these established metrics. On the other hand, metrics like FF-$\alpha$ and FF-$\beta$ show negative correlations, indicating a less favorable alignment with the Human Evaluation score. However, they still provide valuable insights into specific quality aspects that our formula aims to capture, such as temporal and feature consistency.

Our SST-EM final score exhibits the highest correlations (Pearson, Spearman, and Kendall) with the Human Evaluation scores, followed by the Imaging Quality and Aesthetic Quality metrics. This confirms that our final evaluation framework is consistent with human judgments of overall video quality. The individual components - Context Similarity Score, Object Detection Score, and Temporal Consistency Score - each contribute significantly to the overall performance, with Temporal Consistency showing the strongest correlations across all metrics, followed by Object Detection Score.


The high correlation values suggest that our SST-EM evaluation formula performs well not only in alignment with human evaluations but also in comparison to other standard metrics.


\section{Discussion}

We analyze the insights, implications, and limitations of our SST-EM evaluation framework. The results show that SST-EM aligns well with human judgment and outperforms traditional metrics in capturing video editing quality. Our high Pearson correlation (0.962) and perfect Spearman and Kendall correlations (1.000) indicate that our framework ranks video edits similarly to human evaluations, balancing semantic accuracy, temporal consistency, and overall video quality.

While SST-EM aligns well with metrics like Imaging Quality and Aesthetic Quality, traditional metrics like FF-$\alpha$ and FF-$\beta$ show lower correlations, suggesting that they do not fully capture the complexities of video editing, especially in scenes with rapid changes. The lower correlation with the Background Consistency Score highlights its limited ability to assess overall video quality.

\subsection{Insights into SST-EM Metric Components}

Our framework includes Semantic Similarity, Object Detection, Temporal Consistency, and Aesthetic Quality. The high Pearson correlation (0.951) with Imaging Quality and perfect rank agreement with Aesthetic Quality show that our metric captures the semantic and aesthetic aspects effectively. Context similarity from Vision-Language Models (VLMs) ensures semantic accuracy.

The Object Detection and Temporal Consistency components are important but may need refinement for dynamic or cluttered backgrounds and rapid scene transitions. The ViT-based Temporal Consistency Score captures frame coherence but may require further tuning for subtle temporal variations.

\subsection{Comparison to Traditional Metrics}

Compared to CLIP-based metrics, SST-EM offers a more holistic evaluation by considering temporal consistency and object continuity. CLIP metrics excel at semantic alignment but fall short in addressing video-specific challenges like maintaining object consistency. By incorporating human-derived weights, SST-EM provides a more balanced evaluation, better reflecting overall video quality.

In conclusion, SST-EM provides a robust approach to evaluating video editing models. Future work will refine weighting mechanisms and handle complex editing scenarios to optimize performance across diverse tasks.

\section{Conclusion}

We introduce the SST-EM evaluation framework, a novel approach combining context similarity, object detection accuracy, and temporal consistency to address limitations in existing video editing evaluation methodologies. SST-EM captures nuances such as temporal coherence and semantic relevance, offering a multidimensional perspective that aligns moderately with human evaluations.

The key strength of SST-EM lies in its ability to bridge semantic understanding and visual perception, enabling comprehensive evaluations. For future work, we plan to expand the dataset with diverse video samples, incorporate more human evaluation data, and explore deep learning-based weighting models to improve metric adaptability and robustness. We also aim to apply SST-EM to state-of-the-art models, providing insights into their strengths and weaknesses. Ultimately, SST-EM has the potential to standardize video editing evaluation, fostering meaningful comparisons and advancing research in this field.

{\small
\bibliographystyle{ieee_fullname}
\bibliography{egbib}

\begin{thebibliography}{10}\itemsep=-1pt

\bibitem{detectron2}
Allena Venkata~Sai Abhishek and Sonali Kotni.
\newblock Detectron2 object detection \& manipulating images using
  cartoonization.
\newblock {\em Int. J. Eng. Res. Technol.(IJERT)}, 10:1--5, 2021.

\bibitem{zhao-humaneval}
Sharib Ali, Felix Zhou, Adam Bailey, Barbara Braden, James~E East, Xin Lu, and
  Jens Rittscher.
\newblock A deep learning framework for quality assessment and restoration in
  video endoscopy.
\newblock {\em Medical image analysis}, 68:101900, 2021.

\bibitem{text2live}
Omer Bar-Tal, Dolev Ofri-Amar, Rafail Fridman, Yoni Kasten, and Tali Dekel.
\newblock Text2live: Text-driven layered image and video editing.
\newblock In {\em European conference on computer vision}, pages 707--723.
  Springer, 2022.

\bibitem{paligemma}
Lucas Beyer, Andreas Steiner, Andr{\'e}~Susano Pinto, Alexander Kolesnikov,
  Xiao Wang, Daniel Salz, Maxim Neumann, Ibrahim Alabdulmohsin, Michael
  Tschannen, Emanuele Bugliarello, et~al.
\newblock Paligemma: A versatile 3b vlm for transfer.
\newblock {\em arXiv preprint arXiv:2407.07726}, 2024.

\bibitem{chen}
Feng Chen, Zhen Yang, Bohan Zhuang, and Qi Wu.
\newblock Streaming video diffusion: Online video editing with diffusion
  models.
\newblock {\em arXiv preprint arXiv:2405.19726}, 2024.

\bibitem{controlavideo}
Weifeng Chen, Yatai Ji, Jie Wu, Hefeng Wu, Pan Xie, Jiashi Li, Xin Xia, Xuefeng
  Xiao, and Liang Lin.
\newblock Control-a-video: Controllable text-to-video generation with diffusion
  models.
\newblock {\em arXiv preprint arXiv:2305.13840}, 2023.

\bibitem{chen2024editboard}
Yupeng Chen, Penglin Chen, Xiaoyu Zhang, Yixian Huang, and Qian Xie.
\newblock Editboard: Towards a comprehensive evaluation benchmark for
  text-based video editing models.
\newblock {\em arXiv preprint arXiv:2409.09668}, 2024.

\bibitem{clip}
Peng Gao, Shijie Geng, Renrui Zhang, Teli Ma, Rongyao Fang, Yongfeng Zhang,
  Hongsheng Li, and Yu Qiao.
\newblock Clip-adapter: Better vision-language models with feature adapters.
\newblock {\em International Journal of Computer Vision}, 132(2):581--595,
  2024.

\bibitem{tokenflow}
Michal Geyer, Omer Bar-Tal, Shai Bagon, and Tali Dekel.
\newblock Tokenflow: Consistent diffusion features for consistent video
  editing.
\newblock {\em arXiv preprint arXiv:2307.10373}, 2023.

\bibitem{godfrey}
W~Wilfred Godfrey and Abhinav Ratna.
\newblock Enhancing the video editing capabilities of text-to-video generators
  using ddpm inversion.
\newblock In {\em 2023 IEEE International Conference on Computer Vision and
  Machine Intelligence (CVMI)}, pages 1--5. IEEE, 2023.

\bibitem{zhou}
Bo Han, Heqing Zou, Haoyang Li, Guangcong Wang, and Chng~Eng Siong.
\newblock Text-based talking video editing with cascaded conditional diffusion.
\newblock {\em arXiv preprint arXiv:2407.14841}, 2024.

\bibitem{yolov5}
Glenn Jocher, Ayush Chaurasia, Alex Stoken, Jirka Borovec, Yonghye Kwon, Kalen
  Michael, Jiacong Fang, Colin Wong, Zeng Yifu, Diego Montes, et~al.
\newblock ultralytics/yolov5: v6. 2-yolov5 classification models, apple m1,
  reproducibility, clearml and deci. ai integrations.
\newblock {\em Zenodo}, 2022.

\bibitem{vilt}
Wonjae Kim, Bokyung Son, and Ildoo Kim.
\newblock Vilt: Vision-and-language transformer without convolution or region
  supervision.
\newblock In {\em International conference on machine learning}, pages
  5583--5594. PMLR, 2021.

\bibitem{sam}
Alexander Kirillov, Eric Mintun, Nikhila Ravi, Hanzi Mao, Chloe Rolland, Laura
  Gustafson, Tete Xiao, Spencer Whitehead, Alexander~C Berg, Wan-Yen Lo, et~al.
\newblock Segment anything.
\newblock In {\em Proceedings of the IEEE/CVF International Conference on
  Computer Vision}, pages 4015--4026, 2023.

\bibitem{anyv2v}
Max Ku, Cong Wei, Weiming Ren, Huan Yang, and Wenhu Chen.
\newblock Anyv2v: A plug-and-play framework for any video-to-video editing
  tasks.
\newblock {\em arXiv preprint arXiv:2403.14468}, 2024.

\bibitem{lee}
Yao-Chih Lee, Ji-Ze~Genevieve Jang, Yi-Ting Chen, Elizabeth Qiu, and Jia-Bin
  Huang.
\newblock Shape-aware text-driven layered video editing.
\newblock In {\em Proceedings of the IEEE/CVF Conference on Computer Vision and
  Pattern Recognition}, pages 14317--14326, 2023.

\bibitem{blip}
Junnan Li, Dongxu Li, Caiming Xiong, and Steven Hoi.
\newblock Blip: Bootstrapping language-image pre-training for unified
  vision-language understanding and generation.
\newblock In {\em International conference on machine learning}, pages
  12888--12900. PMLR, 2022.

\bibitem{vid2me}
Xirui Li, Chao Ma, Xiaokang Yang, and Ming-Hsuan Yang.
\newblock Vidtome: Video token merging for zero-shot video editing.
\newblock In {\em Proceedings of the IEEE/CVF Conference on Computer Vision and
  Pattern Recognition}, pages 7486--7495, 2024.

\bibitem{groundingDINO}
Shilong Liu, Zhaoyang Zeng, Tianhe Ren, Feng Li, Hao Zhang, Jie Yang, Qing
  Jiang, Chunyuan Li, Jianwei Yang, Hang Su, et~al.
\newblock Grounding dino: Marrying dino with grounded pre-training for open-set
  object detection.
\newblock In {\em European Conference on Computer Vision}, pages 38--55.
  Springer, 2025.

\bibitem{videop2p}
Shaoteng Liu, Yuechen Zhang, Wenbo Li, Zhe Lin, and Jiaya Jia.
\newblock Video-p2p: Video editing with cross-attention control.
\newblock In {\em Proceedings of the IEEE/CVF Conference on Computer Vision and
  Pattern Recognition}, pages 8599--8608, 2024.

\bibitem{trailblazer}
Wan-Duo~Kurt Ma, John~P Lewis, and W~Bastiaan Kleijn.
\newblock Trailblazer: Trajectory control for diffusion-based video generation.
\newblock {\em arXiv preprint arXiv:2401.00896}, 2023.

\bibitem{kumar}
KL~Bhanu Moorthy, Moneish Kumar, Ramanathan Subramanian, and Vineet Gandhi.
\newblock Gazed--gaze-guided cinematic editing of wide-angle monocular video
  recordings.
\newblock In {\em Proceedings of the 2020 CHI Conference on Human Factors in
  Computing Systems}, pages 1--11, 2020.

\bibitem{fatezero}
Chenyang Qi, Xiaodong Cun, Yong Zhang, Chenyang Lei, Xintao Wang, Ying Shan,
  and Qifeng Chen.
\newblock Fatezero: Fusing attentions for zero-shot text-based video editing.
\newblock In {\em Proceedings of the IEEE/CVF International Conference on
  Computer Vision}, pages 15932--15942, 2023.

\bibitem{enhanced}
Lakshmi~Priya Ramisetty, Namrata Patel, Hiep Dang, and Aditya~Singh Parmar.
\newblock Enhanced end-to-end video editing: Adaptive customization of path,
  object, and motion dynamics.

\bibitem{pelechano}
Otger Rogla, Gustavo~A Patow, and Nuria Pelechano.
\newblock Procedural crowd generation for semantically augmented virtual
  cities.
\newblock {\em Computers \& Graphics}, 99:83--99, 2021.

\bibitem{yolov7}
Chien-Yao Wang, Alexey Bochkovskiy, and Hong-Yuan~Mark Liao.
\newblock Yolov7: Trainable bag-of-freebies sets new state-of-the-art for
  real-time object detectors.
\newblock In {\em Proceedings of the IEEE/CVF conference on computer vision and
  pattern recognition}, pages 7464--7475, 2023.

\bibitem{actionclip}
Mengmeng Wang, Jiazheng Xing, Jianbiao Mei, Yong Liu, and Yunliang Jiang.
\newblock Actionclip: Adapting language-image pretrained models for video
  action recognition.
\newblock {\em IEEE Transactions on Neural Networks and Learning Systems},
  2023.

\bibitem{medclip}
Zifeng Wang, Zhenbang Wu, Dinesh Agarwal, and Jimeng Sun.
\newblock Medclip: Contrastive learning from unpaired medical images and text.
\newblock {\em arXiv preprint arXiv:2210.10163}, 2022.

\bibitem{TuneAVideo}
Jay~Zhangjie Wu, Yixiao Ge, Xintao Wang, Stan~Weixian Lei, Yuchao Gu, Yufei
  Shi, Wynne Hsu, Ying Shan, Xiaohu Qie, and Mike~Zheng Shou.
\newblock Tune-a-video: One-shot tuning of image diffusion models for
  text-to-video generation.
\newblock In {\em Proceedings of the IEEE/CVF International Conference on
  Computer Vision}, pages 7623--7633, 2023.

\bibitem{florence}
Bin Xiao, Haiping Wu, Weijian Xu, Xiyang Dai, Houdong Hu, Yumao Lu, Michael
  Zeng, Ce Liu, and Lu Yuan.
\newblock Florence-2: Advancing a unified representation for a variety of
  vision tasks.
\newblock In {\em Proceedings of the IEEE/CVF Conference on Computer Vision and
  Pattern Recognition}, pages 4818--4829, 2024.

\bibitem{xu-HumanEval}
Yiran Xu, Badour AlBahar, and Jia-Bin Huang.
\newblock Temporally consistent semantic video editing.
\newblock In {\em European Conference on Computer Vision}, pages 357--374.
  Springer, 2022.

\bibitem{dong}
Songlin Yang, Wei Wang, Jun Ling, Bo Peng, Xu Tan, and Jing Dong.
\newblock Context-aware talking-head video editing.
\newblock In {\em Proceedings of the 31st ACM International Conference on
  Multimedia}, pages 7718--7727, 2023.

\bibitem{wang-humaneval}
Tianle Zhang, Langtian Ma, Yuchen Yan, Yuchen Zhang, Kai Wang, Yue Yang, Ziyao
  Guo, Wenqi Shao, Yang You, Yu Qiao, et~al.
\newblock Rethinking human evaluation protocol for text-to-video models:
  Enhancing reliability, reproducibility, and practicality.
\newblock {\em arXiv preprint arXiv:2406.08845}, 2024.

\bibitem{zhao2025motiondirector}
Rui Zhao, Yuchao Gu, Jay~Zhangjie Wu, David~Junhao Zhang, Jia-Wei Liu, Weijia
  Wu, Jussi Keppo, and Mike~Zheng Shou.
\newblock Motiondirector: Motion customization of text-to-video diffusion
  models.
\newblock In {\em European Conference on Computer Vision}, pages 273--290.
  Springer, 2025.

\bibitem{detr}
Xizhou Zhu, Weijie Su, Lewei Lu, Bin Li, Xiaogang Wang, and Jifeng Dai.
\newblock Deformable detr: Deformable transformers for end-to-end object
  detection.
\newblock {\em arXiv preprint arXiv:2010.04159}, 2020.

\end{thebibliography}
}

\end{document}